%% file: template.tex
\documentclass[cameraready]{Interspeech}
\title{From Reactive to Proactive: Assessing the Proactivity of Voice Agents via ProVoice-Bench}

\author{Ke}{Xu}
\author{Yuhao}{Wang}
\author[correspondingauthor]{Yu}{Wang}
\address{
    Shanghai Jiao Tong University
}

\email{\{overji1, colane, yuwangsjtu\}@sjtu.edu.cn}


\usepackage{comment}
\usepackage{subcaption}
\usepackage{multirow}
\usepackage{booktabs}
\usepackage{graphicx}
\usepackage{IEEEtrantools}

\begin{document}
\maketitle

\begin{abstract}
    Recent advancements in LLM agents are gradually shifting from reactive, text-based paradigms toward proactive, multimodal interaction. However, existing benchmarks primarily focus on reactive responses, overlooking the complexities of proactive intervention and monitoring. To bridge this gap, we introduce ProVoice-Bench, the first evaluation framework specifically designed for proactive voice agents, featuring four novel tasks. By leveraging a multi-stage data synthesis pipeline, we curate 1,182 high-quality samples for rigorous testing. Our evaluation of state-of-the-art Multimodal LLMs reveals a significant performance gap, particularly regarding over-triggering and reasoning capabilities. These findings highlight the limitations of current models and offer a roadmap for developing more natural, context-aware proactive agents. 
\end{abstract}

\input{sections/1_Introduction}

\input{sections/2_Pipeline}
\input{sections/3_Experiment}
\input{sections/4_Conclusion}
\input{sections/5_GenerativeAIUseDisclosure}

\bibliographystyle{IEEEtran}
\bibliography{mybib}

\end{document}

%% file: sections/1_Introduction.tex
\section{Introduction}

\begin{figure*}[t]
  \centering
    \includegraphics[width=0.95\linewidth]{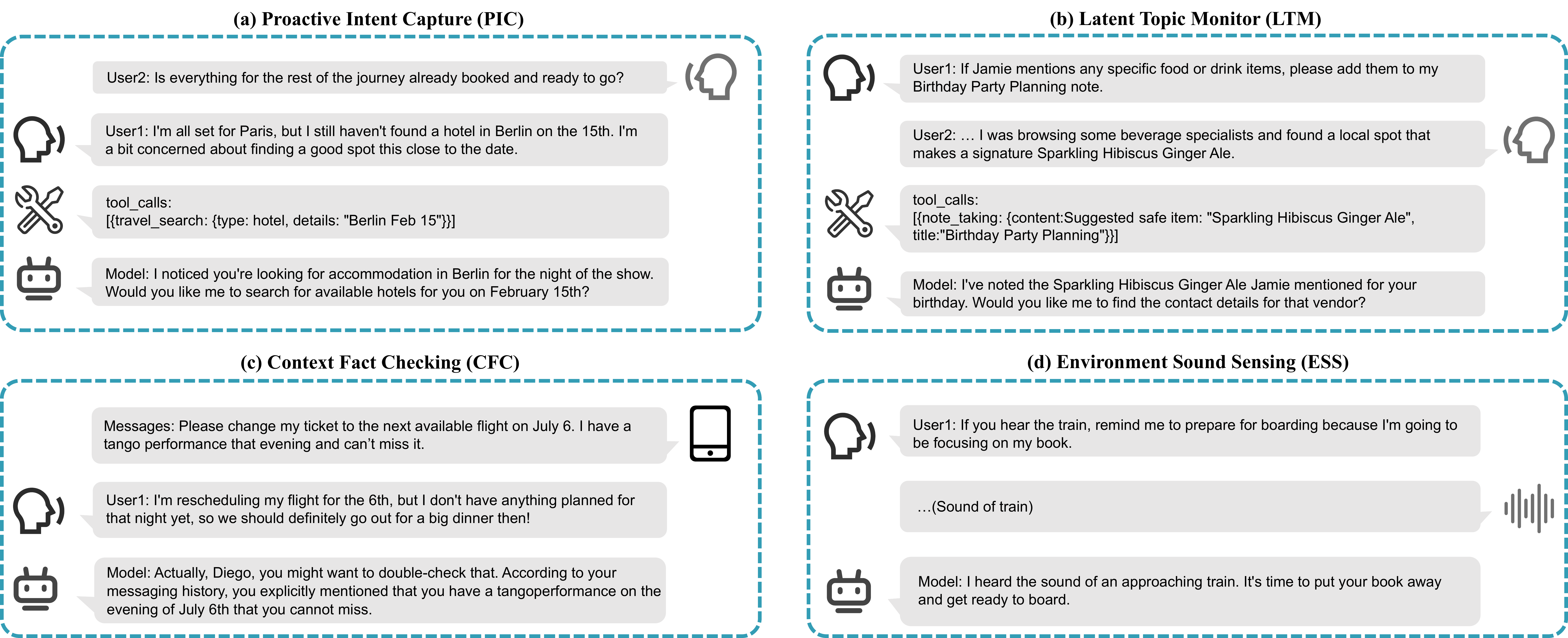}
    \caption{Overview of the four designed tasks in ProVoice-Bench.}
  \label{fig:overview_tasks}
\end{figure*}

With the rapid development of Multimodal Large Language Models (MLLMs)~\cite{comanici2025gemini, wu2025omnigen2explorationadvancedmultimodal, Qwen3-Omni, kimiteam2025kimiaudiotechnicalreport}, we have witnessed the emergence of advanced voice agents capable of perceiving audio and generating natural speech responses via end-to-end models~\cite{zou2026ltsvoiceagentlistenthinkspeakframeworkefficient, jain2025voiceagentbenchvoiceassistantsready}. These voice agents can also execute complex reasoning, planning, and tool use by leveraging their robust textual capabilities~\cite{chen2024voicebenchbenchmarkingllmbasedvoice,chiang2025shankssimultaneoushearingthinking, maben2025auraagentunderstandingreasoning}. While these agents offer a more natural and intelligent interaction experience, they operate primarily in a reactive paradigm, responding only when explicitly instructed. This limitation prevents them from inferring implicit needs or detecting specific trigger points that necessitate intervention\cite{10.1145/3539618.3594250}. In contrast, human communication is inherently proactive; humans possess the ability to flexibly intervene in a conversation based on their continuous assessment of the context and environment~\cite{lu2024proactiveagentshiftingllm}.

In contrast to reactive models, we advocate for the necessity of proactive agents~\cite{wang2025proactivevideoqacomprehensivebenchmarkevaluating, wang2025roboomni}. Such agents should not merely wait for users to instruct them; they should also sense when providing assistance would be helpful. Furthermore, they should have the ability to remain dormant, waiting for specific trigger points—such as a pre-defined topic or a particular environmental sound—before intervening. For instance, a proactive agent might offer assistance upon detecting hesitation in a user's speech, or trigger a reminder when a previously specified alarm sound occurs. Recently, some studies have explored proactive multimodal agents, like ContextAgent~\cite{yang2025contextagentcontextawareproactivellm} and ProAgent~\cite{yang2025proagentharnessingondemandsensory}, but they focus primarily on visual cues and ignore the rich information within the audio modality. Moreover, they often limit proactive interaction to implicit cues, without exploring the paradigm of user-defined trigger points.

To bridge this gap, we introduce ProVoice-Bench, the first evaluation framework specifically designed to assess proactivity in voice agents. The benchmark comprises 1,182 meticulously curated samples across four novel proactive tasks as shown in Figure~\ref{fig:overview_tasks}: (1) \textbf{Proactive Intent Capture (PIC)}, requiring the model to grasp implicit intent within the conversation and initiate tool calls; (2) \textbf{Latent Topic Monitor (LTM)}, where the model monitors dialogues and triggers assistance only upon detecting a user-defined semantic trigger; (3) \textbf{Context Fact Checking(CFC)}, which necessitates interrupting the user when verbal statements contradict digital context records; (4) \textbf{Environment Sound Sensing(ESS)}, where the agent recognizes user-defined acoustic events as cues for intervention. 

Collectively, these four tasks encapsulate the core capabilities required for a robust proactive voice agent. While PIC addresses the traditional challenge of inferring intent from implicit linguistic cues, the remaining tasks introduce pioneering benchmarks for more complex proactive behaviors: LTM and ESS focus on condition-based triggering across dialogues and acoustic environments, and CFC explores knowledge-driven intervention by identifying contradictions between verbal statements and digital contexts. To facilitate rigorous evaluation, we developed a specialized data synthesis pipeline to produce high-fidelity, naturalistic samples. Our subsequent evaluation of state-of-the-art MLLMs uncovers a substantial performance gap, indicating that current models are not yet fully equipped for reliable proactive interaction. The contributions of this work are summarized as follows:
\begin{itemize}
    \item We propose new tasks and a paradigm for proactive agents that integrate audio and digital context, offering a vision for more natural communication with voice agents.
    \item We present the ProVoice-Bench, containing 1,182 carefully selected samples across the four brand-new proactive tasks.
    \item We evaluate several open-source MLLMs on ProVoice-Bench, uncovering a significant performance gap and identifying the current shortcomings of MLLMs in proactive interaction.
\end{itemize}

%% file: sections/2_Pipeline.tex
\section{ProVoice-Bench Construction}

\subsection{Task Overview}
In contrast to existing reactive agents, we propose ProVoice-Bench, a comprehensive evaluation suite comprising a large-scale speech corpus designed for proactive interaction tasks. Within this benchmark, the necessity for interaction is determined by monitoring multimodal signals, including digital context (e.g., mobile application states), as well as implicit cues and explicit triggers within the conversation. We propose four comprehensive proactive tasks as below:

\begin{itemize}
    \item \textbf{Proactive Intent Capture (PIC):} The model infers implicit user intentions from nuanced linguistic cues (e.g., hesitation or prospective action items discussed in dialogue) and proactively initiates tool-call requests while seeking confirmation. Some requests require understanding of the user's digital context for operational accuracy.
    \item \textbf{Latent Topic Monitor (LTM):} The user instructs the model to monitor ambient conversations. The model remains silent, intervening only when a trigger designated by the user is detected in the speech of interlocutors.
    \item \textbf{Environment Sound Sensing (ESS):} The model provides assistance upon detecting specific acoustic events (e.g., alarms), remaining silent until the trigger predefined by the user is recognized.
    \item \textbf{Context Fact Checking (CFC):} When verbal statements contradict digital context on users' mobile phones, the model proactively interrupts to provide factual corrections, ensuring consistency with the user's digital context.
\end{itemize}

We formalize proactive voice agents as: $(T_p, R_p) = \mathcal{A}(\mathcal{C}_a, \mathcal{D}_c)$, where $\mathcal{A}$ denotes the agent model that integrates conversational audio $\mathcal{C}_a$ with the user's digital context $\mathcal{D}_c$. The agent yields a tool-call request $T_p$ and a textual response $R_p$. 

To establish a standardized evaluation framework, each sample in our benchmark is represented by the quintuple $(\mathcal{C}_a, \mathcal{D}_c, \mathcal{S}_c, R_g, T_g)$. Beyond primary inputs $\mathcal{C}_a$ and $\mathcal{D}_c$, we introduce $\mathcal{S}_c$ as the semantic cue, which indicates the conversation context and the key signals that trigger proactive interaction. The tool-call request $T_g$ and textual response $R_g$ serve as ground-truth references to evaluate proactive invocation accuracy and the quality of the agent's response.

\begin{figure*}[t]
  \centering
    \includegraphics[width=\linewidth]{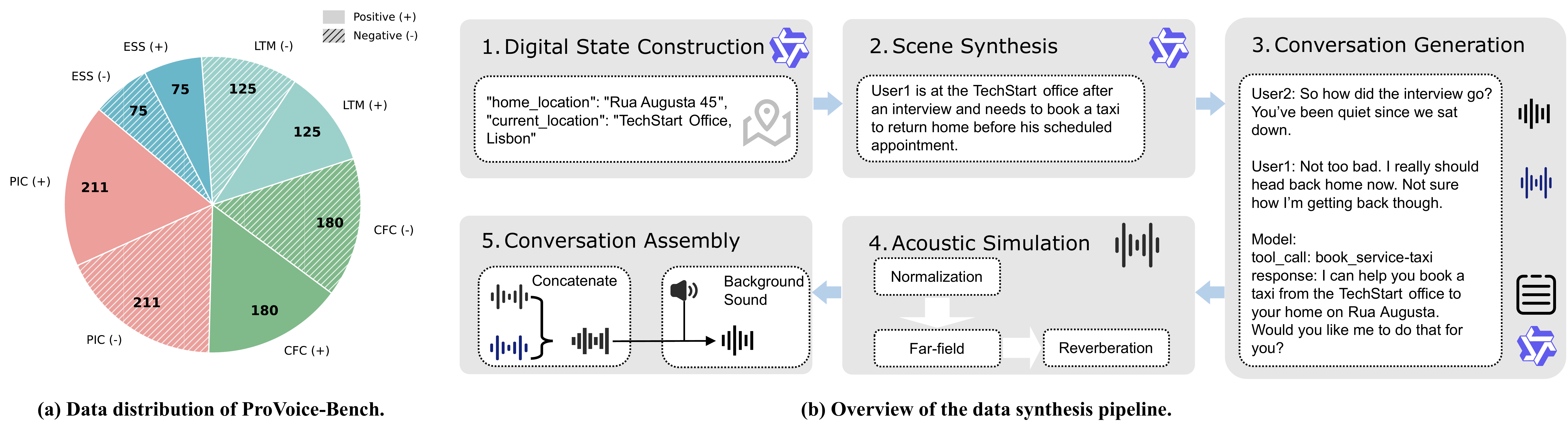}
    \caption{\textbf{ProVoice-Bench data synthesis overview.} (a) Distribution of data across the four tasks in the ProVoice-Bench. (b) Data synthesis pipeline: a multi-stage process for generating semantic cues and corresponding conversational audio.}
  \label{fig:overview}
\end{figure*}

\subsection{Data Construction Pipeline}
To construct the four distinct data categories, we developed a multi-stage pipeline using Large Language Models~(LLMs) and Text-To-Speech~(TTS) models. As shown in Figure \ref{fig:overview}(b), the construction process is divided into the following five phases:
\begin{itemize}
    \item \textbf{Digital State Construction.} The LLM synthesizes mobile application states centered on specific topics, incorporating implicit cues that will be used for proactive interaction tasks.
    \item \textbf{Scene Synthesis.} Based on digital states, we construct high-level scenarios assigned with specific task types (PIC, LTM, ESS, or CFC), requiring the agent to decide between proactive intervention and remaining dormant.
    \item \textbf{Conversation Generation.} High-level scenes are instantiated into detailed scripts, which are subsequently converted into speech segments using state-of-the-art TTS engines with diverse speaker identities and appropriate gender profiles.
    \item \textbf{Acoustic Simulation.} To ensure environmental realism, we post-process the audio streams applying robust acoustic simulation techniques
    \item \textbf{Conversation Assembly.} Speech segments are concatenated using Gaussian-distributed intervals to simulate natural conversation flow, with stochastic environmental noise to simulate realistic acoustic scenes.
\end{itemize}

\subsubsection{Digital State Construction}
To simulate realistic and semantically rich digital contexts, which serve to ground user personas and inform potential future actions, we adopt the application state format from OB2~\cite{veerabadran2025benchmarkingegocentricmultimodalgoal}. The generation process begins by randomly selecting a theme from the dialog-topics dataset~\cite{thatsgroes2025dialogtopics}. Subsequently, we employ Qwen3-Max~\cite{qwen3max} to synthesize fine-grained digital states conditioned on these themes.

These states incorporate implicit cues, such as scheduled appointments (e.g., specific meeting times) or personal constraints (e.g., dietary restrictions or medical conditions). By incorporating these underlying variables, the pipeline provides the necessary contextual foundation for the model to evaluate whether proactive intervention is contextually justified in later stages.

\subsubsection{Scene Synthesis}
During this phase, the LLM analyzes the implicit information within the digital states to construct a natural and unambiguous task-oriented scenario. Each scene is synthesized based on the digital context, a designated task category, and a set of available tools. The resulting output comprises: triggering cues (specific user actions or environmental inputs that elicit proactive model response), a description of the conversational context, and explicit temporal metadata engineered to ensure models can accurately decipher relative timing and event sequences.

\subsubsection{Conversation Generation}
In this stage, we synthesize naturalistic dialogues aligned with the semantic cues and digital contexts established previously. To facilitate the synthesis of lifelike speech, we employ CosyVoice3~\cite{du2025cosyvoice3inthewildspeech} using randomly sampled human speech from seed-tts-eval~\cite{anastassiou2024seedttsfamilyhighqualityversatile} as audio prompts. The selected audio prompts are gender-matched to the characters delineated in the conversation scripts.

We define two distinct roles: User1 denotes the primary user (e.g., equipped with smart earphones or a mobile device) whom the model is tasked to assist, while User2 signifies an interlocutor. For certain scenarios, User2 may be omitted to simulate solo speech or private monologues. For the ESS task, we incorporate specific environmental sound events from the ESC-50 dataset~\cite{piczak2015dataset}, guided by the acoustic parameters specified during the scene generation phase.

\subsubsection{Acoustic Simulation}
To rectify loudness consistency across synthesized outputs derived from audio prompts with varying levels, we normalize all audio signals to a target RMS of \SI{-20}{dBFS}. To simulate far-field effects for the speech of User 2, we apply a \SI{-3}{dB} treble biquad filter at \SI{4}{\kilo\hertz}, accompanied by a \SI{4}{dB} attenuation to model off-axis energy loss. Reverberation is introduced by convolving the signal with a stochastic Room Impulse Response (RIR), synthesized via Gaussian noise modulated by an exponential decay envelope. The final output utilizes a wet/dry ratio of 0.3, approximating the spectral tilt and temporal smearing characteristic of distant speech\cite{barker2018fifthchimespeechseparation}.

\subsubsection{Conversation Assembly}
To simulate natural conversational pacing, inter-turn intervals $\Delta t$ are sampled from a clipped Gaussian distribution $\mathcal{N}(\mu, \sigma)$ within $[t_{\min}, t_{\max}]$. We set parameters $(\mu, \sigma, t_{\min}, t_{\max})$ to $(0.75, 0.35, -1.5, 2.5)s$ for general conversation and $(10.0, 1.66, 2.5, 20.0)s$ for the ESS task to reflect longer trigger latencies. Overlapping speech (negative $\Delta t$) is handled via additive mixing. Finally, to emulate realistic acoustic environments, we incorporate ambient noise randomly selected from CochlScene~\cite{jeong2022cochlsceneacquisitionacousticscene} into the composite speech waveform.

%% file: sections/3_Experiment.tex
\section{Experiment and Result}
\label{sec:experiment_and_result}

\begin{table*}[!t]
\centering
\caption{Proactive interaction performance across various models. (T) denotes thinking models and ``Params'' refers to the number of model parameters in billions (B). Each task is evaluated by Recall (Rec), False Positive Rate (FPR), and Accuracy (Acc).}
\label{tab:audio_benchmark_comprehensive}
\resizebox{0.98\linewidth}{!}{
\begin{tabular}{lcccccccccccccccc}
\toprule
\multirow{2}{*}{\textbf{Model}} & \multirow{2}{*}{\textbf{Params}} & \multicolumn{3}{c}{\textbf{CFC}} & \multicolumn{3}{c}{\textbf{LTM}} & \multicolumn{3}{c}{\textbf{PIC}} & \multicolumn{3}{c}{\textbf{ESS}} & \multicolumn{3}{c}{\textbf{Overall}} \\
\cmidrule(lr){3-5} \cmidrule(lr){6-8} \cmidrule(lr){9-11} \cmidrule(lr){12-14} \cmidrule(lr){15-17}
& & \textbf{Rec $\uparrow$} & \textbf{FPR $\downarrow$} & \textbf{Acc $\uparrow$} & \textbf{Rec $\uparrow$} & \textbf{FPR $\downarrow$} & \textbf{Acc $\uparrow$} & \textbf{Rec $\uparrow$} & \textbf{FPR $\downarrow$} & \textbf{Acc $\uparrow$} & \textbf{Rec $\uparrow$} & \textbf{FPR $\downarrow$} & \textbf{Acc $\uparrow$} & \textbf{Rec $\uparrow$} & \textbf{FPR $\downarrow$} & \textbf{Acc $\uparrow$} \\
\midrule
Mimo-Audio~\cite{coreteam2025mimoaudio}       & 7B  & 0.383 & 0.389 & 0.497 & 0.160 & \textbf{0.096} & 0.532 & 0.848 & 0.289 & 0.780 & 0.293 & \textbf{0.107} & 0.593 & 0.491 & 0.255 & 0.618 \\
Mimo-Audio(T)~\cite{coreteam2025mimoaudio}    & 7B  & 0.633 & 0.078 & 0.778 & 0.680 & 0.504 & 0.588 & 0.838 & 0.237 & 0.800 & 0.824 & 0.533 & 0.644 & 0.740 & 0.283 & 0.729 \\
Qwen3-Omni~\cite{Qwen3-Omni}       & 30B & 0.433 & 0.361 & 0.536 & 0.992 & 0.712 & 0.640 & 0.754 & 0.199 & 0.777 & 0.600 & 0.400 & 0.600 & 0.687 & 0.382 & 0.652 \\
Qwen3-Omni(T)~\cite{Qwen3-Omni}     & 30B & 0.737 & \textbf{0.061} & \textbf{0.838} & 0.920 & 0.256 & \textbf{0.832} & 0.578 & \textbf{0.028} & 0.775 & 0.920 & 0.680 & 0.620 & 0.742 & \textbf{0.169} & 0.787 \\
Step-Audio-R1~\cite{tian2025step}    & 33B & 0.878 & 0.961 & 0.460 & \textbf{1.000} & 0.920 & 0.540 & \textbf{0.995} & 0.934 & 0.531 & 0.840 & 0.360 & \textbf{0.740} & \textbf{0.941} & 0.866 & 0.538 \\
Step-Audio-R1(T)~\cite{tian2025step} & 33B & \textbf{0.739} & 0.083 & 0.828 & \textbf{1.000} & 0.392 & 0.804 & 0.915 & 0.270 & \textbf{0.822} & \textbf{0.893} & 0.680 & 0.607 & 0.876 & 0.291 & \textbf{0.793} \\
Qwen2.5-Omni~\cite{xu2025qwen25omnitechnicalreport}     & 7B  & 0.322 & 0.428 & 0.447 & 0.784 & 0.520 & 0.632 & 0.706 & 0.242 & 0.732 & 0.800 & 0.400 & 0.700 & 0.618 & 0.377 & 0.620 \\
\bottomrule
\end{tabular}
}
\end{table*}

\begin{table*}[!t]
\centering
\caption{Response Accuracy ($R_{acc}$) Performance across Different Categories. (T) denotes thinking models.}
\label{tab:response_accuracy}
\setlength{\tabcolsep}{10pt} % 适当增加列间距使表格更美观
\begin{tabular}{lcccccc}
\toprule
\textbf{Model} & \textbf{Params} & \textbf{CFC} & \textbf{LTM} & \textbf{PIC} & \textbf{ESS} & \textbf{Overall} \\
\midrule
Mimo-Audio~\cite{coreteam2025mimoaudio} & 7B  & 0.377 & 0.477 & 0.587 & 0.560 & 0.496 \\
Mimo-Audio(T)~\cite{coreteam2025mimoaudio} & 7B  & 0.615 & 0.462 & 0.663 & 0.586 & 0.596 \\
Qwen3-Omni~\cite{Qwen3-Omni} & 30B & 0.459 & 0.563 & 0.692 & 0.525 & 0.573 \\
Qwen3-Omni(T)~\cite{Qwen3-Omni} & 30B & \textbf{0.826} & \textbf{0.792} & \textbf{0.734} & 0.617 & \textbf{0.759} \\ 
Step-Audio-R1~\cite{tian2025step} & 33B & 0.229 & 0.443 & 0.395 & \textbf{0.693} & 0.393 \\
Step-Audio-R1(T)~\cite{tian2025step} & 33B & 0.806 & 0.741 & 0.722 & 0.587 & 0.734 \\
Qwen2.5-Omni~\cite{xu2025qwen25omnitechnicalreport} & 7B  & 0.328 & 0.483 & 0.594 & 0.555 & 0.484 \\
\bottomrule
\end{tabular}
\end{table*}

\subsection{ProVoice-Bench Overview} 
As illustrated in Figure \ref{fig:overview}(a), to evaluate proactive interaction capabilities across diverse scenarios, we construct the ProVoice-Bench. This benchmark comprises 1,182 meticulously curated multimodal samples, balanced with both positive and negative instances.

\subsection{Metrics} 
To comprehensively assess proactive voice agents, we introduce two categories of metrics focusing on \textit{interaction decision-making} and \textit{response quality}.

\begin{itemize}
    \item \textbf{Proactive Interaction Prediction:} We assess the model's proficiency in determining intervention necessity. Specifically, we employ \textbf{Accuracy~(Acc)} to measure overall binary decision performance, the \textbf{False Positive Rate~(FPR)} to quantify the frequency of unnecessary interventions in the absence of valid triggers, and \textbf{Recall~(Rec)} to evaluate the model's sensitivity in identifying actual interaction triggers.

    \item \textbf{Response Accuracy ($R_{acc}$):} Beyond the initial trigger detection, the correctness of the subsequent action is critical. We define a comprehensive score $S_i$ for each instance $i$ by integrating tool-calling precision and semantic response alignment:
    
    \begin{equation}
    \begin{split}
        S_i = \ & \frac{\mathcal{J}(\mathcal{S}_{c,i}, T_{\text{p}, i}, T_{\text{g}, i}) + \mathcal{J}(\mathcal{S}_{c,i}, R_{\text{p}, i}, R_{\text{g}, i})}{2} \\
        & \cdot \mathbb{I}(\text{pred}_i = \text{gt}_i)
    \end{split}
    \label{eq:instance_score}
    \end{equation}
    
    Where:
    \begin{itemize}
        \item $\mathcal{J}(\mathcal{S}_{c,i}, \cdot, \cdot) \in \{0, 0.5, 1.0\}$ denotes the \textbf{LLM-as-a-Judge} function. Taking the semantic cue $\mathcal{S}_c$ as the evaluation context, the judge evaluates the functional rationality and alignment of the predicted output against the ground truth. Here we use Qwen3-80B~\cite{qwen3technicalreport} as the judge model.
        \item $T_{\text{p}, i}$ and $T_{\text{g}, i}$ represent the predicted and ground-truth tool-call sequences, respectively.
        \item $R_{\text{p}, i}$ and $R_{\text{g}, i}$ denote the predicted and ground-truth textual responses.
        \item $\mathbb{I}(\cdot)$ is the \textbf{indicator function}, which outputs 1 if the model correctly identifies the interaction trigger (i.e., whether an intervention is required), and 0 otherwise.
    \end{itemize}

    The final metric, \textbf{Response Accuracy}, is calculated as the arithmetic mean across the entire benchmark:
    \begin{equation}
        R_{acc} = \frac{1}{N} \sum_{i=1}^{N} S_i
        \label{eq:total_res_acc}
    \end{equation}
    where $N$ denotes the total number of samples in the ProVoice-Bench.
\end{itemize}

\subsection{Results on ProVoice-Bench} 
The experimental results are summarized in Table \ref{tab:audio_benchmark_comprehensive} and Table \ref{tab:response_accuracy}. Based on these results, we draw the following observations:

\noindent \textbf{Propensity for Over-triggering.} The results reveal a widespread tendency toward over-triggering. This is particularly evident in LTM tasks, where most models tend to respond regardless of whether actual trigger points exist in the conversation. Similarly, in CFC tasks, models frequently fail to recognize when a conversation proceeds without any violation of the digital context.

\noindent \textbf{Chain-of-Thought (CoT) Enhances Analysis-Intensive Tasks.} Experimental evidence suggests that CoT~\cite{shao2024deepseekmathpushinglimitsmathematical} significantly improves performance in CFC, LTM, and PIC tasks. These tasks require models to analyze implicit cues embedded in either the digital context or user speech. Furthermore, CoT helps bridge the gap between the decision to interact and the execution of an appropriate response.

\noindent \textbf{Discrepancy Between Decision-to-Speak and Task Execution.} The results highlight a fundamental gap between ``knowing when to speak'' and ``knowing what to do.'' Current models frequently suffer from semantic drift during interactions or generate hallucinated tool calls. This underscores the urgent need for future research to prioritize context awareness and environment sensing in MLLMs.
\begin{figure}[h]
  \centering
  \includegraphics[width=0.98\linewidth]{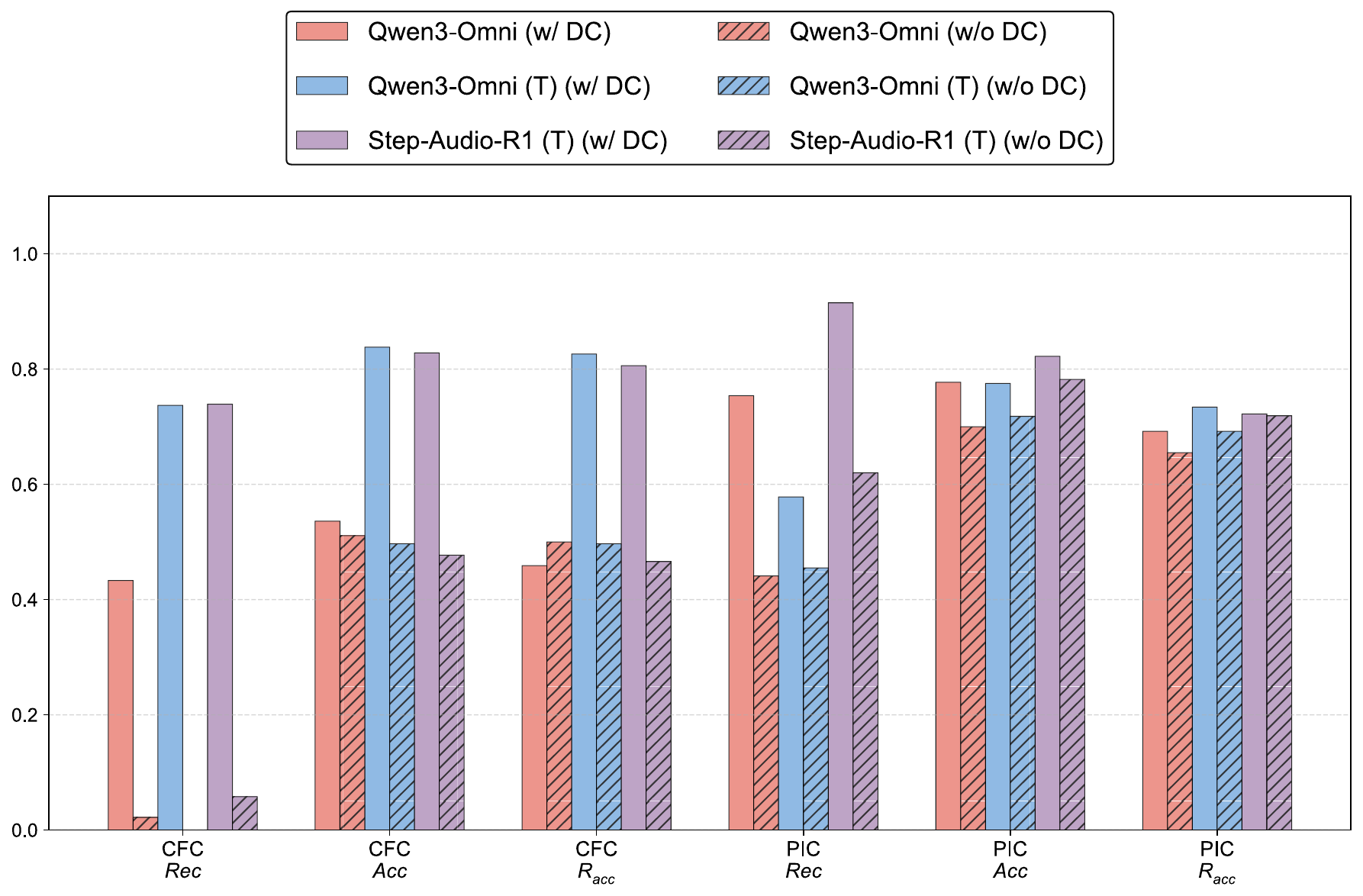}
  \caption{Experiments comparing model performance with (w/ DC) and without (w/o DC) Digital Context on ProVoice-Bench.}
  \label{fig:digital_context_remove_study}
\end{figure}

\subsection{Impact of Digital Context}
We conduct experiments to investigate the impact of digital context by omitting it from the benchmark. As illustrated in Figure \ref{fig:digital_context_remove_study}, the absence of digital context leads to a substantial decline in Recall for the CFC task, which relies on such information to determine interaction timing. Similarly, this removal results in decreased Accuracy and Recall in PIC tasks, where digital context is essential for accurately inferring users' implicit intentions.

%% file: sections/4_Conclusion.tex
\section{Conclusion}
We presented ProVoice-Bench, the first evaluation suite for proactive audio agents, featuring 1,182 high-quality samples across four novel tasks. By integrating digital context with audio input, our benchmark shifts the agent paradigm from reactive responses to context-aware, proactive interaction. Experimental results on state-of-the-art MLLMs reveal a significant performance gap, particularly in over-triggering and bridging the ``decision-to-execution'' divide. These findings underscore the necessity of enhanced context awareness for future multimodal agents. We hope ProVoice-Bench serves as a catalyst for developing more natural and autonomous proactive voice assistants.

%% file: sections/5_GenerativeAIUseDisclosure.tex
\section{Generative AI Use Disclosure}
The authors used Gemini solely for editing and polishing the language and grammar of this manuscript to improve its readability. All scientific analysis, interpretation of results, and the writing of the manuscript were performed by the human authors, who remain fully responsible for the work and its integrity.